# First Principle Approach to Modeling of Small Scale Helicopter

A. Budiyono*, T. Sudiyanto† and H. Lesmana†

*Center for Unmanned System Studies
Institut Teknologi Bandung, Indonesia
e-mail: agus.budiyono@ae.itb.ac.id

†Aeronautics and Astronautics Department
Institut Teknologi Bandung, Indonesia

**Abstract**

The establishment of global helicopter linear model is very precious and useful for the design of the linear control laws, since it is never afforded in the published literatures. In the first principle approach, the mathematical model was developed using basic helicopter theory accounting for particular characteristic of the miniature helicopter. No formal system identification procedures are required for the proposed model structure. The relevant published literatures however did not present the linear models required for the design of linear control laws. The paper presents a step by step development of linear model for small scale helicopter based on first-principle approach. Beyond the previous work in literatures, the calculation of the stability derivatives is presented in detail. A computer program is used to solve the equilibrium conditions and then calculate the change in aerodynamics forces and moments due to the change in each degree of freedom and control input. The detail derivation allows the comprehensive analysis of relative dominance of vehicle states and input variables to force and moment components. Hence it facilitates the development of minimum complexity small scale helicopter dynamics model.

## 1   Introduction

The dynamics of rotary wing vehicles are substantially more complex than those of its fixed-wing counterparts. For full-scale helicopters the aerodynamics analysis was presented in literatures such as [1] and [2]. The treatment for stability and control were presented in [3] and [4]. The rigid body dynamics approach standard in the analysis of aircraft dynamics as in references [5], [6] and [7] is insufficient to capture key dynamics of a rotorcraft vehicle. The rotorcraft dynamics are characteristically typified by the coupled rotor-fuselage dynamics. The small scale rotorcraft dynamics is further characterized by the existence of stabilizer bar and active-yaw damping system to ease the pilot workload.

The rotor head of small scale helicopter is significantly more rigid than that of its full-scale counterpart. This not only allows for larger rotor control moment but also alleviation of second order effects typically found in full-scale helicopter dynamics model. The majority developed model for RUAVs are based on frequency-domain identification. It has been demonstrated that the relatively low order model developed using this approach is sufficient to describe the helicopter dynamics around trim conditions. It must be noted however that the accuracy of this approach decreases with the presence of feedback which is needed for example to stabilize the helicopter in hover. The method is also unreliable to describe low-frequency modes, primarily due to pilot feedback [8]. Other linear model development was given by [9] by using time-domain identification. The system identification approach requires experimental input-output data collected from the flight tests of the vehicle. The flying test-bed must be outfitted with adequate instruments to measure both state and control variables.

The paper presents an analytical development of linear model for small scale helicopter based on first principle approach. Beyond the previous work in [8], the calculation of stability and control derivatives to construct the linear model is presented in detail. The analytical model derivation allows the comprehensive analysis of relative dominance of vehicle states and input variables to force and moment components. And hence it facilitates the development of minimum complexity small scale helicopter dynamics model that differs from that of its full-scale counterpart. In the presented simplified model, the engine drive-train dynamics and inflow dynamics are not necessary to be taken into consideration. The additional rotor degrees of freedom for coning and lead-lag can be omitted for small scale helicopters. It is demonstrated analytically that the dynamics of small scale helicopter is dominated by the strong moments produced by the highly rigid rotor. The dominant rotor forces and moments largely overshadow the effects of complex interactions between the rotor wake and fuselage or tail. This tendency substantially reduces the need for complicated models of second-degree effects typically found in the literature on full-scale helicopters.

The presented approach is not limited to specific trim conditions like hover or forward flights and therefore can be used to develop a global model of small scale rotorcraft vehicle to the purpose of practical control design. The contributions of the thesis are two-folds: an analytical development of linear model of small-scale rotorcraft vehicle based on first-principle approach; and the development of novel algebraic approach for control design. To date, to the author knowledge, the works represent distinct treatment not available in the literatures.

## 2 Dynamics of small scale helicopter

The approach to helicopter modeling can be in general divided into two distinct methods. The first approach is known as first principle modeling based on direct physical understanding of forces and moments balance of the vehicle. The challenge of this approach is the complexity of the mathematical model involved along with the need for rigorous validation. The second method based on system identification basically arises from the difficulty of the former approach. The frequency domain identification starts with the estimation of frequency response from flight data recorder from an instrumented flight-test vehicle. The parameterized dynamic model can then be developed in the form of a linear state-space model using physical insight and frequency-response analysis. The identification can also be conducted in time-domain.

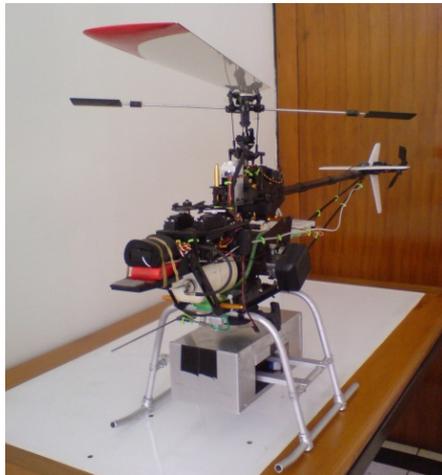

**Figure 1:** Instrumented X-Cell 60 helicopter

The author argues that, any modeling should start from adequate basis in first-principle. In practice, the above two methods can be used hand in hand in developing an accurate small scale rotorcraft vehicle model for the purpose of control design.

### 2.1 Small-scale helicopter parameters
The X-Cell .60 SE small scale helicopter's parameters are used in developing mathematical model. The mini helicopter as shown in **Figure 1:** is characterized by a hinge-less rotor with a diameter of 0.775 m and mass of 8 kg. The X-Cell blades both for main and tail rotors use symmetric airfoils. **Table 1** shows basic X-cell 60 helicopter's specifications.

| Parameter | | Description |
|---|---|---|
| $m$ | 8.2 kg | Helicopter mass |
| $I_{xx}$ | 0.18 kg m$^2$ | Rolling moment of inertia |
| $I_{yy}$ | 0.34 kg m$^2$ | Pitching moment of inertia |
| $I_{zz}$ | 0.28 kg m$^2$ | Yawing moment of inertia |
| $R_{MR}$ | 0.775 m | Main rotor radius |
| $c_{MR}$ | 0.058 m | Main rotor chord |
| $R_{TR}$ | 0.13 m | Tail rotor radius |
| $c_{TR}$ | 0.029 m | Tail rotor chord |
| $S_{VF}$ | 0.012 m$^2$ | Effective vertical fin area |
| $S_{HF}$ | 0.01 m$^2$ | Horizontal fin area |
| $h_{MR}$ | 0.235 m | Main rotor hub height above c.g. |
| $l_{TR}$ | 0.91 m | Tail rotor hub location behind c.g. |
| $h_{TR}$ | 0.08 m | Tail rotor height above c.g. |
| $\Omega_{nom}$ | 167 rad/sec | Nominal main rotor speed |

**Table 1: Basic XCell heli parameters**

### 2.2 Euler-Newton equations of motion
Rigid body equations of motion are typically used for modeling the dynamics of fixed wing aircraft. This model is typified by the use of linear stability derivatives, which are a linearized form of the rigid-body dynamics, providing important analysis of the vehicle's dynamics, stability and control. This approach has a limited application for rotorcraft vehicles particularly for the design of high-bandwidth control design and rigorous handling-qualities analysis. However, it must be noted that the linear stability derivative models from rigid-body dynamics can be a good starting point for a development of more accurate rotorcraft model.

The motion of a vehicle in three-dimensional space can be represented by the position of the center of mass and the Euler angles for the vehicle rotation with respect to the inertial frame of reference. The Euler-Newton equations are derived from the law of conservation of linear and angular momentum. Assuming that vehicle mass $m$ and inertial tensor $\bar{I}$, the equations are given by:

$$m \left. \frac{d\bar{V}}{dt} \right|_{\bar{I}} = \bar{F} \qquad (1)$$
$$\bar{I} \left. \frac{d\bar{\omega}}{dt} \right|_{\bar{I}} = \bar{M}$$

where $\bar{F} = \begin{bmatrix} X & Y & Z \end{bmatrix}^T$ is the vector of external forces acting on the helicopter center of gravity and $\bar{M} = \begin{bmatrix} L & M & N \end{bmatrix}^T$ is the vector of external moments. For helicopter, the external forces and moments consists of forces generated by the main rotor, tail rotor; aerodynamics forces from fuselage, horizontal fin and vertical fin and gravitational force. For computational convenience, the Euler-Newton equations describing the rigid-body dynamics of the helicopter is then represented with respect to body coordinate system by using the kinematic principles of moving coordinate frame of reference as the following:

$$m\dot{\bar{V}} + m(\bar{\omega} \times \bar{V}) = \bar{F}$$
$$I\dot{\bar{\omega}} + (\bar{\omega} \times I\bar{\omega}) = \bar{M} \quad (2)$$

Here the vector $\bar{V} = [u \ v \ w]^T$ and $\bar{\omega} = [p \ q \ r]^T$ are the fuselage velocities and angular rates in the body coordinate system, respectively. For the helicopter moving in six degrees of freedom, the above equations produce six differential equations describing the vehicle's translational motion and angular motion about its three reference axes.

$$\sum X = m(\dot{u} - rv + qw) + mg\sin\theta \quad (3)$$
$$\sum Y = m(ru + \dot{v} - pw) - mg\sin\phi\cos\theta \quad (4)$$
$$\sum Z = m(-qu + pv - \dot{w}) - mg\cos\phi\cos\theta \quad (5)$$
$$\sum L = I_{xx}\dot{p} - (I_{yy} - I_{zz})qr \quad (6)$$
$$\sum M = I_{yy}\dot{q} - (I_{zz} - I_{xx})pr \quad (7)$$
$$\sum N = I_{zz}\dot{r} - (I_{xx} - I_{yy})pq \quad (8)$$
$$\dot{\phi} = p + (q\sin\phi + r\cos\phi)\tan\theta \quad (9)$$
$$\dot{\theta} = q\cos\phi - r\sin\phi \quad (10)$$
$$\dot{\psi} = (q\sin\phi + r\cos\phi)\sec\theta \quad (11)$$

### 2.2.1 Main rotor, and tail rotor

The main rotor thrust is calculated by the following expressions.

$$T_{MR} = \rho(\Omega R)_{MR}^2 (\pi R^2)_{MR} C_{TMR} \quad (12)$$

$$C_{TMR} = \frac{1}{2}a_{MR}\sigma_{MR}\left[\frac{1}{2}(\mu_{zMR} - \lambda_{0MR}) + \left(\frac{1}{3} + \frac{1}{2}\mu_{MR}^2\right)\theta_{0MR}\right] \quad (13)$$

$$\lambda_{0MR} \equiv \frac{w_{iMR}}{(\Omega R)_{MR}} = \frac{C_{TMR}}{2\eta_w\sqrt{\mu_{MR}^2 + (\lambda_{0MR} - \mu_{zMR})^2}} \quad (14)$$

$$\mu_{MR} \equiv \frac{\sqrt{u_a^2 + v_a^2}}{(\Omega R)_{MR}} \quad \mu_{zMR} \equiv \frac{w_a}{(\Omega R)_{MR}}$$

These expressions are solved by iteration.

The torque is given by the following expressions.

$$Q_{MR} = \rho(\Omega R)_{MR}^2 (\pi R^2)_{MR} R_{MR} C_{QMR} \quad (15)$$

$$C_{QMR} = \frac{1}{8}\sigma_{MR}\left(1 + \frac{7}{3}\mu_{MR}^2\right)C_{D_0 MR} + (\lambda_{0MR} - \mu_{zMR})C_{TMR} \quad (16)$$

Tail rotor thrust and torque are calculated by the same manner as those of main rotor, by including the main rotor wake, $K_\lambda$, and vertical fin's blockage factor, $f_t$.

$$K_\lambda = 1.5\frac{\left(\frac{u_a}{w_{iMR} - w_a}\right) - g_i}{g_f - g_i} \quad (17)$$

$$f_t = 1 - \frac{3}{4}\frac{S_{VF}}{\pi R_{TR}^2}$$

where

$$g_i = \frac{l_{TR} - R_{MR} - R_{TR}}{h_{TR}}; \ g_f = \frac{l_{TR} - R_{MR} + R_{TR}}{h_{TR}} \quad (18)$$

The main rotor flapping motions are given by the following expressions.

$$\tau_e\dot{a}_{1s} = -a_{1s} + \frac{\partial a_{1s}}{\partial \mu_{MR}}\frac{u_a}{(\Omega R)_{MR}} + \frac{\partial a_{1s}}{\partial \mu_{zMR}}\frac{w_a}{(\Omega R)_{MR}} \cdots$$
$$\cdots -\tau_e q + A_{\delta_{Long}}\delta_{Long} \quad (19)$$

$$\tau_e\dot{b}_{1s} = -b_{1s} - \frac{\partial b_{1s}}{\partial \mu_{MR}}\frac{v_a}{(\Omega R)_{MR}} - \tau_e p + B_{\delta_{Lat}}\delta_{Lat} \quad (20)$$

### 2.2.2 Fuselage, horizontal fin, vertical fin

Fuselage, horizontal fin and vertical fin are subject to drag force generated by relative airspeed, which comes from the motion of helicopter's body relative to air and rotor's induced wind.

## 2.3 Trim condition

For helicopter, there are in general two distinct trim conditions: hover and forward flight. Other conditions include steady turns and helices, which are excluded in this paper.

### 2.3.1 Hover

The hover condition is characterized by zero velocities and angular rates.

$$\bar{V}_0 = [0 \ 0 \ 0]^T$$
$$\bar{\omega}_0 = [0 \ 0 \ 0]^T \quad (21)$$

This expression is used to solve (3), (4), (5), (6), (7), and (8) for the trim value at hover.

$$T_{MR}|_{hov} = 81.616 \text{ Newton} \quad C_{TMR}|_{hov} = 0.002256$$
$$Q_{MR}|_{hov} = 6.247 \text{ N}\cdot\text{m} \quad C_{QMR}|_{hov} = 0.0002228 \quad (22)$$
$$w_{iMR}|_{hov} = 4.582 \text{ m/s}$$
$$\theta_{0MR}|_{hov} = 0.1047 \text{ rad} = 6.001°$$

$$T_{TR}|_{hov} = 6.8656 \text{ Newton} \quad C_{TTR}|_{hov} = 0.01329$$
$$Q_{TR\,hov} = 0.1268 \text{ N}\cdot\text{m} \quad C_{QTR\,hov} = 0.001568 \quad (23)$$
$$v_{iTR}|_{hov} = 8.693 \text{ m/s}$$
$$\theta_{0TR}|_{hov} = 0.2412 \text{ rad} = 13.82°$$

$$a_{1s}|_{hov} = 0.0014258 \text{ rad} = 0.0817° \quad \delta_{Long}|_{hov} = 0.0003395 \text{ rad} = 0.01945° \quad (24)$$
$$b_{1s}|_{hov} = 0.0074866 \text{ rad} = 0.4290° \quad \delta_{Lat}|_{hov} = 0.001783 \text{ rad} = 0.1021°$$

$$\theta|_{hov} = -0.0014471 \text{ rad} = -0.0829°$$
$$\phi|_{hov} = 0.077643 \text{ rad} = 4.4486° \quad (25)$$

### 2.3.2 Forward flight

The trim conditions for forward flight are derived for the following flight condition.

$$u = 16.5557 \text{ m/s} \quad p = 0 \text{ rad/s}$$
$$v = 0.7456 \text{ m/s} \quad q = 0 \text{ rad/s} \quad (26)$$
$$w = 0.2585 \text{ m/s} \quad r = 0 \text{ rad/s}$$

The flight condition represents a nearly straight level flight. Step-by-step derivations are elaborated similar to the case for hover.

$$T_{MR}|_{ff} = 82.145 \text{ Newton} \quad C_{TMR}|_{ff} = 0.002270$$
$$Q_{MR}|_{ff} = 4.660 \text{ N}\cdot\text{m} \quad C_{QMR}|_{ff} = 0.000166 \quad (27)$$
$$w_{iMR}|_{ff} = 1.272 \text{ m/s}$$
$$\theta_{0MR}|_{ff} = 0.0622 \text{ rad} = 3.564°$$

$$T_{TR}|_{ff} = 5.1032 \text{ Newton} \quad C_{TTR}|_{ff} = 0.00988$$
$$Q_{TR\,ff} = 0.0571 \text{ N}\cdot\text{m} \quad C_{QTR}|_{ff} = 0.000706 \quad (28)$$
$$v_{iTR}|_{ff} = 3.336 \text{ m/s}$$
$$\theta_{0TR}|_{ff} = 0.1171 \text{ rad} = 6.71°$$

$$a_{1s}|_{ff} = 0.00547335 \text{ rad} = 0.3136° \quad \delta_{Long}|_{ff} = 0.00039302 \text{ rad} = 0.0225° \quad (29)$$
$$b_{1s}|_{ff} = 0.00558899 \text{ rad} = 0.3202° \quad \delta_{Lat}|_{ff} = 0.001613 \text{ rad} = 0.0924°$$

$$\theta\big|_{cr} = -0.203044 \text{ rad} = -11.6336° \quad (30)$$
$$\phi\big|_{cr} = 0.0790896 \text{ rad} = 4.5315°$$

## 2.4 Linearized equations of motion

The vehicle equations of motion in general can be represented by a nonlinear vector differential equation as follows.

$$\dot{\bar{x}} = \bar{f}(\bar{x}, \bar{u}) \quad (31)$$

where $\bar{x}$ is the vehicle state vector and $\bar{u}$ is the control input vector. This nonlinear differential equation of motion is linearized around an equilibrium state $\bar{x}_0$:

$$\delta\dot{\bar{x}} = \left(\frac{\partial \bar{f}}{\partial \bar{x}}\right)_{\bar{x}_0, \bar{u}_0} \delta\bar{x} + \left(\frac{\partial \bar{f}}{\partial \bar{u}}\right)_{\bar{x}_0, \bar{u}_0} \delta\bar{u} \quad (32)$$

(32) uses the linear perturbations of the state and input vectors of the vehicle. This principle of small perturbations is then applied to each of the expression for forces and moments of the helicopter: (3), (4), (5), (6), (7), (8), (9), (10), and (11).

$$d\dot{u} = \frac{1}{m}\sum dX + rdv - qdw - wdq + vdr - g\sin\theta d\theta \quad (33)$$

$$d\dot{v} = \frac{1}{m}\sum dY - rdu + pdw + wdp - udr \cdots \quad (34)$$
$$\cdots + g\cos\phi\cos\theta d\phi - g\sin\phi\sin\theta d\theta$$

$$d\dot{w} = -\frac{1}{m}\sum dZ - qdv + pdv + vdp - udq \cdots \quad (35)$$
$$\cdots + g\cos\phi\sin\phi d\phi + g\cos\phi\sin\theta d\theta$$

$$d\dot{p} = \frac{1}{I_{xx}}\sum dL + \frac{(I_{yy} - I_{zz})}{I_{xx}}rdq + \frac{(I_{yy} - I_{zz})}{I_{xx}}qdr \quad (36)$$

$$d\dot{q} = \frac{1}{I_{yy}}\sum dM + \frac{(I_{zz} - I_{xx})}{I_{yy}}rdp + \frac{(I_{zz} - I_{xx})}{I_{yy}}pdr \quad (37)$$

$$d\dot{r} = \frac{1}{I_{zz}}\sum dN + \frac{(I_{xx} - I_{yy})}{I_{zz}}qdp + \frac{(I_{xx} - I_{yy})}{I_{zz}}pdq \quad (38)$$

$$d\dot{\phi} = dp + \sin\phi\tan\theta dq + \cos\phi\tan\theta dr \cdots$$
$$\cdots + (q\cos\phi - r\sin\phi)\tan\theta d\phi \cdots$$
$$\cdots + (q\sin\phi + r\cos\phi)\sec\theta d\theta \quad (39)$$

$$d\dot{\theta} = \cos\phi dq - \sin\phi dr - (q\sin\phi + r\cos\phi)d\phi \quad (40)$$

$$d\dot{\psi} = \sin\phi\sec\theta dq + \cos\phi\sec\theta dr \cdots$$
$$\cdots + (q\cos\phi - r\sin\phi)\sec\theta d\phi \cdots$$
$$\cdots + (q\sin\phi + r\cos\phi)\tan\theta\sec\theta d\theta \quad (41)$$

where

$$\sum d(\bullet) = \begin{pmatrix} \frac{\partial d(\bullet)}{\partial du}du + \frac{\partial d(\bullet)}{\partial dv}dv + \frac{\partial d(\bullet)}{\partial dw}dw \cdots \\ \cdots + \frac{\partial d(\bullet)}{\partial dp}dp + \frac{\partial d(\bullet)}{\partial dq}dq + \frac{\partial d(\bullet)}{\partial dr}dr \cdots \\ \cdots + \frac{\partial d(\bullet)}{\partial d\phi}d\phi + \frac{\partial d(\bullet)}{\partial d\theta}d\theta + \frac{\partial d(\bullet)}{\partial d\psi}d\psi \cdots \\ \cdots + \frac{\partial d(\bullet)}{\partial d\delta_{Coll}}d\delta_{Coll} + \frac{\partial d(\bullet)}{\partial d\delta_{Ped}}d\delta_{Ped} \cdots \\ \cdots + \frac{\partial d(\bullet)}{\partial d\delta_{Lat}}d\delta_{Lat} + \frac{\partial d(\bullet)}{\partial d\delta_{Long}}d\delta_{Long} \end{pmatrix} \quad (42)$$

$$(\bullet) = X, Y, Z, L, M, N$$

(33), (34), (35), (36), (37), (38), (39), (40), and (41) can be expressed in compact form.

$$\dot{\bar{x}} = \mathbf{A}\bar{x} + \mathbf{B}\bar{u} \quad (43)$$

where

$$\mathbf{A} = \begin{bmatrix} X_u & X_v & X_w & X_p & X_q & X_r & X_\phi & X_\theta & X_\psi \\ Y_u & Y_v & Y_w & Y_p & Y_q & Y_r & Y_\phi & Y_\theta & Y_\psi \\ Z_u & Z_v & Z_w & Z_p & Z_q & Z_r & Z_\phi & Z_\theta & Z_\psi \\ L_u & L_v & L_w & L_p & L_q & L_r & L_\phi & L_\theta & L_\psi \\ M_u & M_v & M_w & M_p & M_q & M_r & M_\phi & M_\theta & M_\psi \\ N_u & N_v & N_w & N_p & N_q & N_r & N_\phi & N_\theta & N_\psi \\ 0 & 0 & 0 & \Phi_p & \Phi_q & \Phi_r & \Phi_\phi & \Phi_\theta & 0 \\ 0 & 0 & 0 & 0 & \Theta_q & \Theta_r & \Theta_\phi & 0 & 0 \\ 0 & 0 & 0 & 0 & \Psi_q & \Psi_r & \Psi_\phi & \Psi_\theta & 0 \end{bmatrix} \quad (44)$$

$$\mathbf{B} = \begin{bmatrix} X_{\delta_{Coll}} & X_{\delta_{Ped}} & X_{\delta_{Lat}} & X_{\delta_{Long}} \\ Y_{\delta_{Coll}} & Y_{\delta_{Ped}} & Y_{\delta_{Lat}} & Y_{\delta_{Long}} \\ Z_{\delta_{Coll}} & Z_{\delta_{Ped}} & Z_{\delta_{Lat}} & Z_{\delta_{Long}} \\ L_{\delta_{Coll}} & L_{\delta_{Ped}} & L_{\delta_{Lat}} & L_{\delta_{Long}} \\ M_{\delta_{Coll}} & M_{\delta_{Ped}} & M_{\delta_{Lat}} & M_{\delta_{Long}} \\ N_{\delta_{Coll}} & N_{\delta_{Ped}} & N_{\delta_{Lat}} & N_{\delta_{Long}} \\ \Phi_{\delta_{Coll}} & \Phi_{\delta_{Ped}} & \Phi_{\delta_{Lat}} & \Phi_{\delta_{Long}} \\ \Theta_{\delta_{Coll}} & \Theta_{\delta_{Ped}} & \Theta_{\delta_{Lat}} & \Theta_{\delta_{Long}} \\ \Psi_{\delta_{Coll}} & \Psi_{\delta_{Ped}} & \Psi_{\delta_{Lat}} & \Psi_{\delta_{Long}} \end{bmatrix} \quad (45)$$

$$\bar{x} = \begin{bmatrix} du & dv & dw & dp & dq & dr & d\phi & d\theta & d\psi \end{bmatrix}^T \quad (46)$$

$$\bar{u} = \begin{bmatrix} d\delta_{Coll} & d\delta_{Ped} & d\delta_{Lat} & d\delta_{Long} \end{bmatrix}^T \quad (47)$$

The elements of **A** and **B** are obtained by partially derived the total force and moment perturbation toward each degree of freedom plus re-expressing the kinematic relation in term of each degree of freedom.

$$\begin{aligned} X_u &= \frac{1}{m}\frac{\partial dX}{\partial du}, \; X_v = \left(\frac{1}{m}\frac{\partial dX}{\partial dv} + r\right), \cdots \\ \cdots, Y_v &= \frac{1}{m}\frac{\partial dY}{\partial dv}, \; Y_w = \left(\frac{1}{m}\frac{\partial dY}{\partial dw} + p\right), \cdots \\ \cdots, Z_w &= -\frac{1}{m}\frac{\partial dZ}{\partial dw}, \; Z_p = \left(-\frac{1}{m}\frac{\partial dZ}{\partial dp} + v\right), \cdots \\ \cdots, L_p &= \frac{1}{I_{xx}}\frac{\partial dL}{\partial dp}, \; L_q = \frac{1}{I_{xx}}\left(\frac{\partial dL}{\partial dq} + (I_{yy} - I_{zz})r\right), \cdots \\ \cdots, M_q &= \frac{1}{I_{yy}}\frac{\partial dM}{\partial dq}, \; M_r = \frac{1}{I_{yy}}\left(\frac{\partial dM}{\partial dr} + (I_{zz} - I_{xx})p\right), \cdots \\ \cdots, N_r &= \frac{1}{I_{zz}}\frac{\partial dN}{\partial dr}, \; N_\phi = \frac{1}{I_{zz}}\frac{\partial dN}{\partial d\phi}, \cdots \\ \Phi_p &= 1, \; \Phi_q = \sin\phi\tan\theta, \; \Phi_r = \cos\phi\tan\theta, \cdots \\ \cdots, \Phi_\phi &= (q\cos\phi - r\sin\phi)\tan\theta, \; \Phi_\theta = (q\sin\phi + r\cos\phi)\sec\theta. \\ \Theta_q &= \cos\phi, \; \Theta_r = -\sin\phi, \; \Theta_\phi = -(q\sin\phi + r\cos\phi). \\ \Psi_q &= \sin\phi\sec\theta, \; \Psi_r = \cos\phi\sec\theta, \cdots \\ \cdots, \Psi_\phi &= (q\cos\phi - r\sin\phi)\sec\theta, \; \Psi_\theta = (q\sin\phi + r\cos\phi)\tan\theta\sec\theta. \end{aligned} \quad (48)$$

## 2.5 The stability derivatives

The force and moment derivatives of each helicopter component can be obtained by analytical calculation, except for those of main rotor's and tail rotor's, which require iterative calculation. A computer program is used to solve the equilibrium condition and to calculate the change in aerodynamic forces and moments due to the change in each degree of freedom and control input. The following subsections summarized the values of the derivatives.

### 2.5.1 Main rotor derivatives

To assign the values to the derivatives, all parameters are plotted as function of the states and inputs. The value of the derivative can be calculated as the gradient of the curve evaluated at the trim condition.

#### 2.5.1.1 Flapping derivatives

For flapping derivatives, the plots are given in the following figures (only those with respect to $u$), followed by their values (omitting zero vectors).

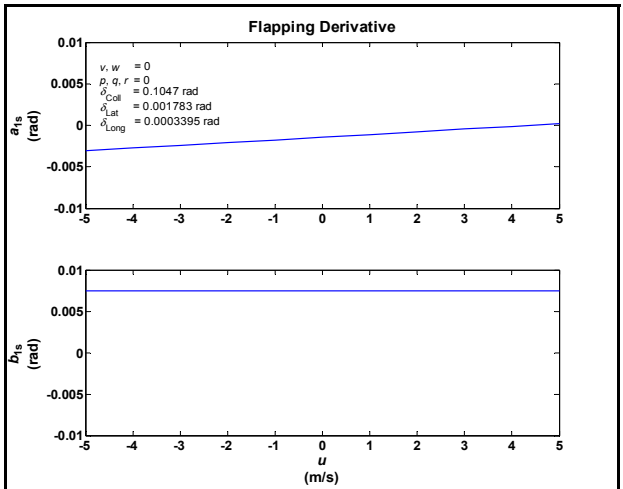

**Figure 2:** Flapping angle with respect to $u$ during hover

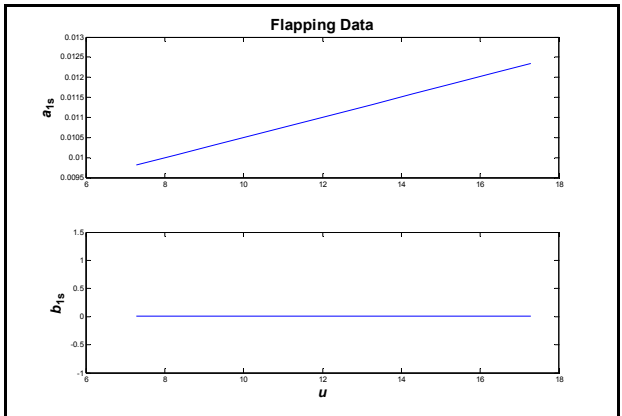

**Figure 3:** Flapping angle with respect to $u$ during forward flight

| Derivatives | Hover | Forward Flight | Unit |
|---|---|---|---|
| $\left[\dfrac{\partial a_{1s}}{\partial u} \dfrac{\partial a_{1s}}{\partial v} \dfrac{\partial a_{1s}}{\partial w}\right]^T$ | $\begin{bmatrix}0.000322\\0\\0\end{bmatrix}$ | $\begin{bmatrix}0.000232\\0\\0.000318\end{bmatrix}$ | rad·s/m |
| $\left[\dfrac{\partial a_{1s}}{\partial p} \dfrac{\partial a_{1s}}{\partial q} \dfrac{\partial a_{1s}}{\partial r}\right]^T$ | $\begin{bmatrix}0\\-0.1\\0\end{bmatrix}$ | $\begin{bmatrix}0\\-0.1\\0\end{bmatrix}$ | s |
| $\left[\dfrac{\partial a_{1s}}{\partial \delta_{Coll}} \dfrac{\partial a_{1s}}{\partial \delta_{Ped}}\right]^T$ | $\begin{bmatrix}0\\0\end{bmatrix}$ | $\begin{bmatrix}0.068218\\0\end{bmatrix}$ | 1 |
| $\left[\dfrac{\partial a_{1s}}{\partial \delta_{Lat}} \dfrac{\partial a_{1s}}{\partial \delta_{Long}}\right]^T$ | $\begin{bmatrix}0\\-4.2\end{bmatrix}$ | $\begin{bmatrix}0\\4.1988\end{bmatrix}$ | 1 |

**Table 2: Longitudinal flapping derivatives**

| Derivatives | Hover | Forward Flight | Unit |
|---|---|---|---|
| $\left[\dfrac{\partial b_{1s}}{\partial u} \dfrac{\partial b_{1s}}{\partial v} \dfrac{\partial b_{1s}}{\partial w}\right]^T$ | $\begin{bmatrix}0\\0.000322\\0\end{bmatrix}$ | $\begin{bmatrix}0\\0.000259\\0\end{bmatrix}$ | rad·s/m |
| $\left[\dfrac{\partial b_{1s}}{\partial p} \dfrac{\partial b_{1s}}{\partial q} \dfrac{\partial b_{1s}}{\partial r}\right]^T$ | $\begin{bmatrix}-0.1\\0\\0\end{bmatrix}$ | $\begin{bmatrix}-0.1\\0\\0\end{bmatrix}$ | s |
| $\left[\dfrac{\partial b_{1s}}{\partial \delta_{Coll}} \dfrac{\partial b_{1s}}{\partial \delta_{Ped}}\right]^T$ | $\begin{bmatrix}0\\0\end{bmatrix}$ | $\begin{bmatrix}0.003073\\0\end{bmatrix}$ | 1 |
| $\left[\dfrac{\partial b_{1s}}{\partial \delta_{Lat}} \dfrac{\partial b_{1s}}{\partial \delta_{Long}}\right]^T$ | $\begin{bmatrix}4.2\\0\end{bmatrix}$ | $\begin{bmatrix}4.1988\\0\end{bmatrix}$ | 1 |

**Table 3: Lateral flapping derivatives**

#### 2.5.1.2 Thrust derivatives

The thrust derivatives plots are given in the following figures (only those with respect to $u$), followed by their values.

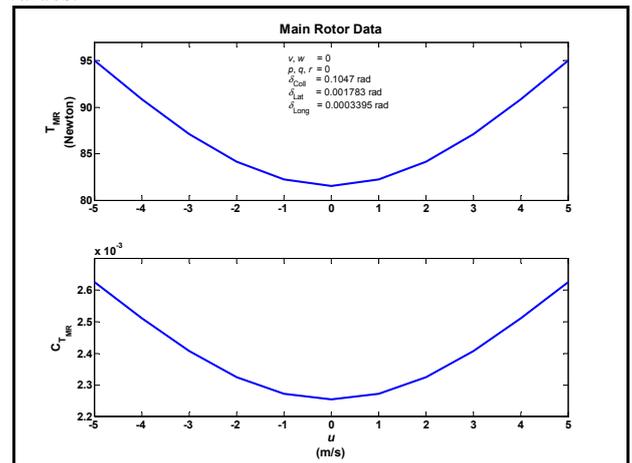

**Figure 4:** Main rotor thrust with respect to $u$ during hover

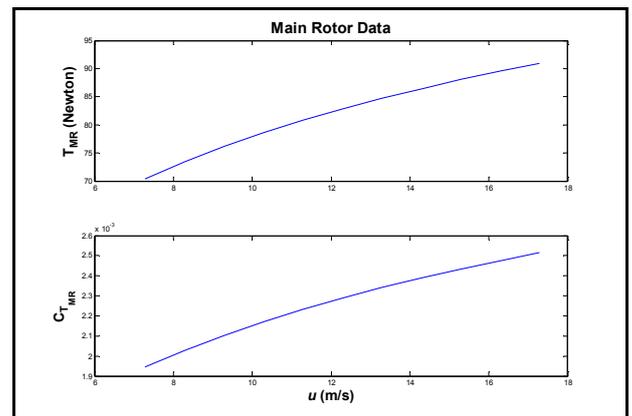

**Figure 5:** Main rotor thrust with respect to $u$ during forward flight

| Derivatives | Hover | Forward Flight | Unit |
|---|---|---|---|
| $\left[\dfrac{\partial T_{MR}}{\partial u}\ \dfrac{\partial T_{MR}}{\partial v}\ \dfrac{\partial T_{MR}}{\partial w}\right]^T$ | $\begin{bmatrix}0\\0\\0.000335\end{bmatrix}$ | $\begin{bmatrix}1.315574\\0.057189\\14.2086\end{bmatrix}$ | N·s/m |
| $\left[\dfrac{\partial T_{MR}}{\partial \delta_{Coll}}\ \dfrac{\partial T_{MR}}{\partial \delta_{Ped}}\right]^T$ | $\begin{bmatrix}1043.6\\0\end{bmatrix}$ | $\begin{bmatrix}1262.5898\\0\end{bmatrix}$ | N/rad |
| $\left[\dfrac{\partial T_{MR}}{\partial \delta_{Lat}}\ \dfrac{\partial T_{MR}}{\partial \delta_{Long}}\right]^T$ | $\begin{bmatrix}0\\0\end{bmatrix}$ | $\begin{bmatrix}0\\0\end{bmatrix}$ | N/rad |

**Table 4: Main rotor thrust derivatives**

### 2.5.1.3 Torque derivatives

The torque derivatives plots are given in the following figures (only those with respect to *u*), followed by their values.

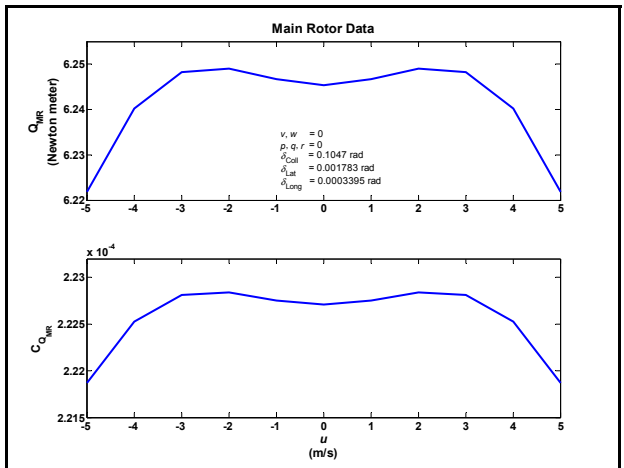

**Figure 6:** Main rotor torque with respect to *u* during hover

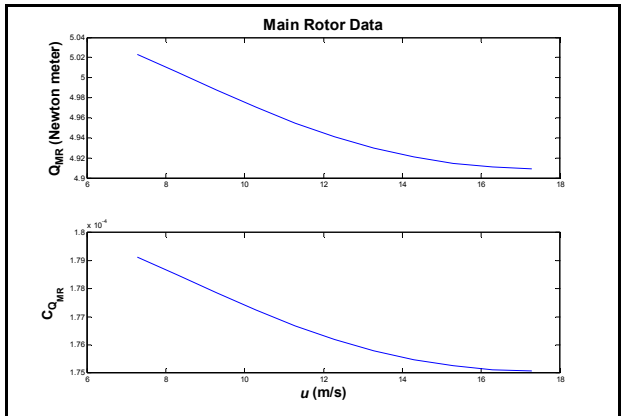

**Figure 7:** Main rotor torque with respect to *u* during forward flight

| Derivatives | Hover | Forward Flight | Unit |
|---|---|---|---|
| $\left[\dfrac{\partial Q_{MR}}{\partial u}\ \dfrac{\partial Q_{MR}}{\partial v}\ \dfrac{\partial Q_{MR}}{\partial w}\right]^T$ | $\begin{bmatrix}0\\0\\0\end{bmatrix}$ | $\begin{bmatrix}-0.001048\\-0.000043\\-0.292165\end{bmatrix}$ | N·s/m |
| $\left[\dfrac{\partial Q_{MR}}{\partial \delta_{Coll}}\ \dfrac{\partial Q_{MR}}{\partial \delta_{Ped}}\right]^T$ | $\begin{bmatrix}42.935\\0\end{bmatrix}$ | $\begin{bmatrix}17.094606\\0\end{bmatrix}$ | N/rad |

**Table 5: Main rotor torque derivatives**

### 2.5.1.4 Induced Velocity derivatives

The torque derivatives plots are given in the following figures (only those with respect to *u*), followed by their values.

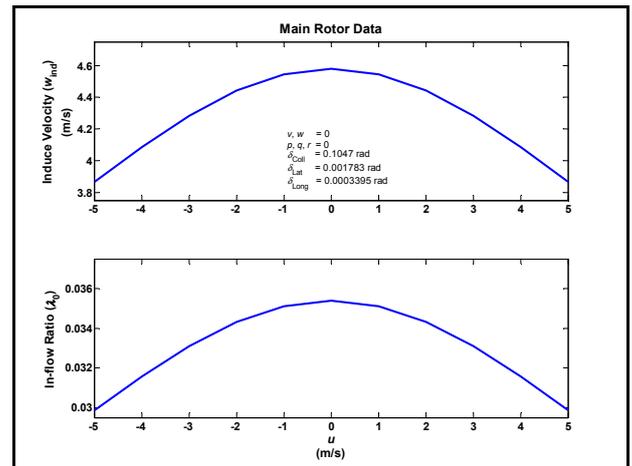

**Figure 8:** Main rotor induced velocity with respect to *u* during hover

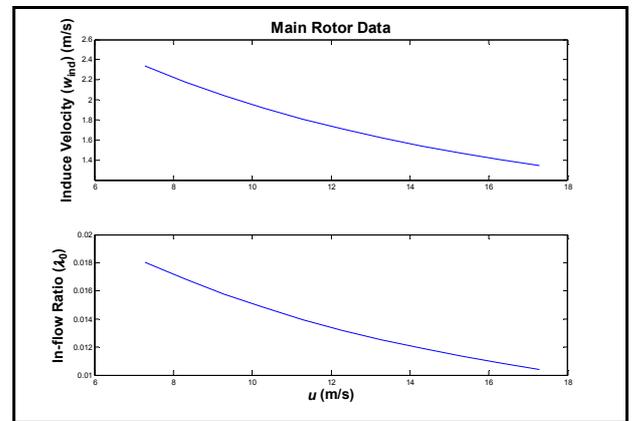

**Figure 9:** Main rotor induced velocity with respect to *u* during forward flight

| Derivatives | Hover | Forward Flight | Unit |
|---|---|---|---|
| $\left[\dfrac{\partial w_{iMR}}{\partial u}\ \dfrac{\partial w_{iMR}}{\partial v}\ \dfrac{\partial w_{iMR}}{\partial w}\right]^T$ | $\begin{bmatrix}0\\0\\-0.000018\end{bmatrix}$ | $\begin{bmatrix}-0.055941\\-0.002430\\0.223859\end{bmatrix}$ | 1 |
| $\left[\dfrac{\partial w_{iMR}}{\partial \delta_{Coll}}\ \dfrac{\partial w_{iMR}}{\partial \delta_{Ped}}\right]^T$ | $\begin{bmatrix}29.301\\0\end{bmatrix}$ | $\begin{bmatrix}19.468431\\0\end{bmatrix}$ | m/rad·s |
| $\left[\dfrac{\partial w_{iMR}}{\partial \delta_{Lat}}\ \dfrac{\partial w_{iMR}}{\partial \delta_{Long}}\right]^T$ | $\begin{bmatrix}0\\0\end{bmatrix}$ | $\begin{bmatrix}0\\0\end{bmatrix}$ | m/rad·s |

**Table 6: Main rotor induced velocity derivatives**

### 2.5.1.5 Main rotor force and moment derivatives

Provided data obtained in **2.5.1.1** to **2.5.1.4**, the main rotor force and moment derivatives can then be calculated by the following expression.

$$\dfrac{\partial X_{MR}}{\partial u} = -\left(a_{1s0}\dfrac{\partial T_{MR}}{\partial u} + T_{MR0}\dfrac{\partial a_{1s}}{\partial u}\right) \quad (49)$$

$$\vdots$$

$$\dfrac{\partial X_{MR}}{\partial \delta_{Long}} = -\left(a_{1s0}\dfrac{\partial T_{MR}}{\delta_{Long}} + T_{MR0}\dfrac{\partial a_{1s}}{\delta_{Long}}\right)$$

$$\dfrac{\partial Y_{MR}}{\partial u} = b_{1s0}\dfrac{\partial T_{MR}}{\partial u} + T_{MR0}\dfrac{\partial b_{1s}}{\partial u} \quad (50)$$

$$\vdots$$

$$\dfrac{\partial Y_{MR}}{\partial \delta_{Long}} = b_{1s0}\dfrac{\partial T_{MR}}{\partial \delta_{Long}} + T_{MR0}\dfrac{\partial b_{1s}}{\partial \delta_{Long}}$$

$$\frac{\partial Z_{MR}}{\partial u} = -\frac{\partial T_{MR}}{\partial u}$$
$$\vdots$$
$$\frac{\partial Z_{MR}}{\partial \delta_{Long}} = -\frac{\partial T_{MR}}{\partial \delta_{Long}}$$
(51)

$$\frac{\partial L_{MR}}{\partial u} = \left(K_\beta + T_{MR0} h_{MR} \cos b_{1s0}\right)\frac{\partial b_{1s}}{\partial u} + \frac{\partial T_{MR}}{\partial u} h_{MR} \sin b_{1s0}$$
$$\vdots$$
$$\frac{\partial L_{MR}}{\partial \delta_{Long}} = \left(K_\beta + T_{MR0} h_{MR} \cos b_{10}\right)\frac{\partial b_{1s}}{\partial \delta_{Long}} + \frac{\partial T_{MR}}{\partial \delta_{Long}} h_{MR} \sin b_{1s0}$$
(52)

$$\frac{\partial M_{MR}}{\partial u} = \left(K_\beta + T_{MR0} h_{MR} \cos a_{1s0}\right)\frac{\partial a_{1s}}{\partial u} + \frac{\partial T_{MR}}{\partial u} h_{MR} \sin a_{1s0}$$
$$\vdots$$
$$\frac{\partial M_{MR}}{\partial \delta_{Long}} = \left(K_\beta + T_{MR0} h_{MR} \cos a_{1s0}\right)\frac{\partial a_{1s}}{\partial \delta_{Long}} + h_{MR} \sin a_{1s0} \frac{\partial T_{MR}}{\partial \delta_{Long}}$$
(53)

$$\frac{\partial N_{MR}}{\partial u} = -\frac{\partial Q_{MR}}{\partial u}$$
$$\vdots$$
$$\frac{\partial N_{MR}}{\partial \delta_{Long}} = -\frac{\partial Q_{MR}}{\partial \delta_{Long}}$$
(54)

### 2.5.2 Tail rotor derivatives

Tail rotor thrust and torque are calculated by the same manner as those of main rotor.

#### 2.5.2.1 Tail rotor thrust, torque and induced velocity derivatives

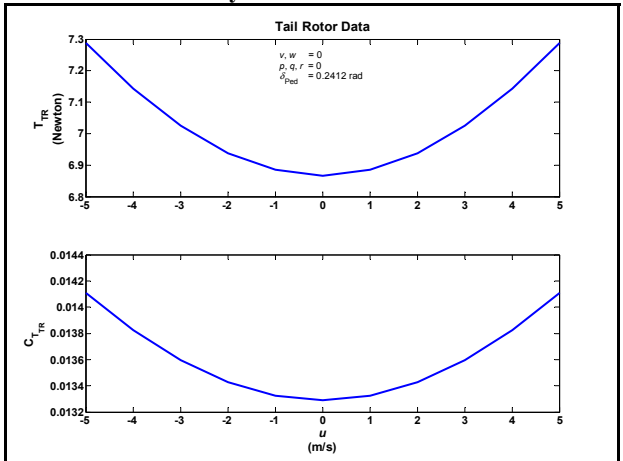

**Figure 10: Tail rotor thrust with respect to $u$ during hover**

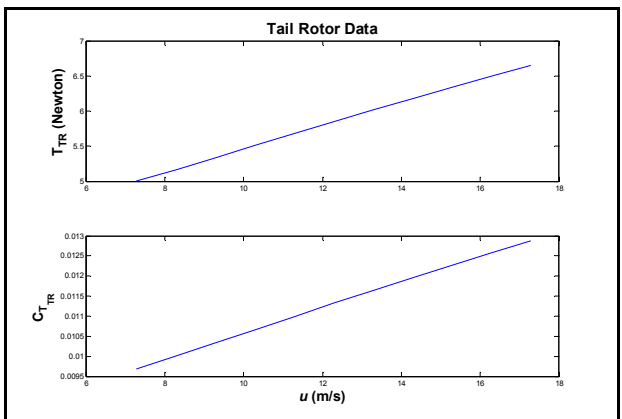

**Figure 11: Tail rotor thrust with respect to $u$ during forward flight**

| Derivatives | Hover | Forward Flight | Unit |
|---|---|---|---|
| $\left[\frac{\partial T_{TR}}{\partial u} \frac{\partial T_{TR}}{\partial v} \frac{\partial T_{TR}}{\partial w}\right]^T$ | $\begin{bmatrix}0\\0\\0\end{bmatrix}$ | $\begin{bmatrix}0.13098\\0.558553\\-0.013237\end{bmatrix}$ | N·s/m |
| $\left[\frac{\partial T_{TR}}{\partial p} \frac{\partial T_{TR}}{\partial q} \frac{\partial T_{TR}}{\partial r}\right]^T$ | $\begin{bmatrix}0\\0\\0\end{bmatrix}$ | $\begin{bmatrix}0.044684\\-0.018719\\-0.508284\end{bmatrix}$ | N·s/rad |
| $\left[\frac{\partial T_{TR}}{\partial \delta_{Coll}} \frac{\partial T_{TR}}{\partial \delta_{Ped}}\right]^T$ | $\begin{bmatrix}0\\38.849\end{bmatrix}$ | $\begin{bmatrix}0\\40.389359\end{bmatrix}$ | N/rad |
| $\left[\frac{\partial T_{TR}}{\partial \delta_{Lat}} \frac{\partial T_{TR}}{\partial \delta_{Long}}\right]^T$ | $\begin{bmatrix}0\\0\end{bmatrix}$ | $\begin{bmatrix}0\\0\end{bmatrix}$ | N/rad |

**Table 7: Tail rotor thrust derivatives**

| Derivatives | Hover | Forward Flight | Unit |
|---|---|---|---|
| $\left[\frac{\partial Q_{TR}}{\partial u} \frac{\partial Q_{TR}}{\partial v} \frac{\partial Q_{TR}}{\partial w}\right]^T$ | $\begin{bmatrix}0\\0\\0\end{bmatrix}$ | $\begin{bmatrix}-0.000053\\-0.0026\\0.000005\end{bmatrix}$ | N·s/m |
| $\left[\frac{\partial Q_{TR}}{\partial p} \frac{\partial Q_{TR}}{\partial q} \frac{\partial Q_{TR}}{\partial r}\right]^T$ | $\begin{bmatrix}0\\0\\0\end{bmatrix}$ | $\begin{bmatrix}-0.000021\\0.000005\\0.002391\end{bmatrix}$ | N·s/rad |
| $\left[\frac{\partial Q_{TR}}{\partial \delta_{Coll}} \frac{\partial Q_{TR}}{\partial \delta_{Ped}}\right]^T$ | $\begin{bmatrix}0\\0.78383\end{bmatrix}$ | $\begin{bmatrix}0\\0.362617\end{bmatrix}$ | N/rad |

**Table 8: Tail rotor torque derivatives**

| Derivatives | Hover | Forward Flight | Unit |
|---|---|---|---|
| $\left[\frac{\partial v_{iTR}}{\partial u} \frac{\partial v_{iTR}}{\partial v} \frac{\partial v_{iTR}}{\partial w}\right]^T$ | $\begin{bmatrix}0\\0\\0\end{bmatrix}$ | $\begin{bmatrix}-0.106211\\0.383865\\0.010723\end{bmatrix}$ | 1 |
| $\left[\frac{\partial v_{iTR}}{\partial p} \frac{\partial v_{iTR}}{\partial q} \frac{\partial v_{iTR}}{\partial r}\right]^T$ | $\begin{bmatrix}0\\0\\0\end{bmatrix}$ | $\begin{bmatrix}0.030709\\0.009618\\-0.349317\end{bmatrix}$ | m/rad |
| $\left[\frac{\partial v_{iTR}}{\partial \delta_{Coll}} \frac{\partial v_{iTR}}{\partial \delta_{Ped}}\right]^T$ | $\begin{bmatrix}0\\24.592\end{bmatrix}$ | $\begin{bmatrix}0\\25.628783\end{bmatrix}$ | m/rad·s |

**Table 9: Tail rotor induced velocity derivatives**

#### 2.5.2.2 Tail rotor force and moment derivatives

Provided data obtained in **2.5.2.1**, the main rotor force and moment derivatives can then be calculated by the following expression.

$$X_{TR} = 0 \quad \frac{\partial Y_{TR}}{\partial u} = -\frac{\partial T_{TR}}{\partial u} \quad Z_{TR} = 0 \quad (55)$$
$$\vdots$$
$$\frac{\partial Y_{TR}}{\partial \delta_{Long}} = -\frac{\partial T_{TR}}{\partial \delta_{Long}}$$

$$\frac{\partial L_{TR}}{\partial u} = -\frac{\partial T_{TR}}{\partial u} h_{TR} \quad \frac{\partial M_{TR}}{\partial u} = -\frac{\partial Q_{TR}}{\partial u} \quad \frac{\partial N_{TR}}{\partial u} = \frac{\partial T_{TR}}{\partial u} l_{TR} \quad (56)$$
$$\vdots$$
$$\frac{\partial L_{TR}}{\partial \delta_{Long}} = -\frac{\partial T_{TR}}{\partial \delta_{Long}} h_{TR} \quad \frac{\partial M_{TR}}{\partial \delta_{Long}} = -\frac{\partial Q_{TR}}{\partial \delta_{Long}} \quad \frac{\partial N_{TR}}{\partial \delta_{Long}} = \frac{\partial T_{TR}}{\partial \delta_{Long}} l_{TR}$$

## 3 Stability Analysis

According to ([4], 1964), we can assume that the cross-coupling dynamics is insignificant. Therefore, rearranging **(44)** and **(45)** for consecutive longitudinal and lateral-directional states respectively (including main rotor flapping dynamic as augmented states), we have these expressions for **A** and **B**.

$$\mathbf{A} = \begin{bmatrix} \mathbf{A}_{\text{long-ver}} & \mathbf{A}_{(\text{lat-dir})2(\text{long-ver})} \\ \mathbf{A}_{(\text{long-ver})2(\text{lat-dir})} & \mathbf{A}_{\text{lat-dir}} \end{bmatrix} \quad (57)$$

$$\mathbf{B} = \begin{bmatrix} \mathbf{B}_{\text{long-ver}} & \mathbf{B}_{(\text{lat-dir})2(\text{long-ver})} \\ \mathbf{B}_{(\text{long-ver})2(\text{lat-dir})} & \mathbf{B}_{\text{lat-dir}} \end{bmatrix}$$

where

$$\mathbf{A}_{\text{long-ver}} = \begin{bmatrix} X_u & X_w & X_q & X_\theta & X_{a_{1s}} \\ Z_u & Z_w & Z_q & Z_\theta & Z_{a_{1s}} \\ M_u & M_w & M_q & M_\theta & M_{a_{1s}} \\ \Theta_u & \Theta_w & \Theta_q & \Theta_\theta & \Theta_{a_{1s}} \\ \frac{1}{\tau_e}A_u & \frac{1}{\tau_e}A_w & \frac{1}{\tau_e}A_q & \frac{1}{\tau_e}A_\theta & \frac{1}{\tau_e}A_{a_{1s}} \end{bmatrix} \mathbf{B}_{\text{long-ver}} = \begin{bmatrix} X_{\delta_{\text{Coll}}} & X_{\delta_{\text{Long}}} \\ Z_{\delta_{\text{Coll}}} & Z_{\delta_{\text{Long}}} \\ M_{\delta_{\text{Coll}}} & M_{\delta_{\text{Long}}} \\ \Theta_{\delta_{\text{Coll}}} & \Theta_{\delta_{\text{Long}}} \\ \frac{1}{\tau_e}A_{\delta_{\text{Coll}}} & \frac{1}{\tau_e}A_{\delta_{\text{Long}}} \end{bmatrix} \quad (58)$$

$$\mathbf{A}_{\text{lat-dir}} = \begin{bmatrix} Y_v & Y_p & Y_r & Y_\phi & Y_{b_{1s}} \\ L_v & L_p & L_r & L_\phi & L_{b_{1s}} \\ N_v & N_p & N_r & N_\phi & N_{b_{1s}} \\ \Phi_v & \Phi_p & \Phi_r & \Phi_\phi & \Phi_{b_{1s}} \\ \frac{1}{\tau_e}B_v & \frac{1}{\tau_e}B_p & \frac{1}{\tau_e}B_r & \frac{1}{\tau_e}B_\phi & \frac{1}{\tau_e}B_{b_{1s}} \end{bmatrix} \mathbf{B}_{\text{lat-dir}} = \begin{bmatrix} Y_{\delta_{\text{Ped}}} & Y_{\delta_{\text{Lat}}} \\ L_{\delta_{\text{Ped}}} & L_{\delta_{\text{Lat}}} \\ N_{\delta_{\text{Ped}}} & N_{\delta_{\text{Lat}}} \\ \Phi_{\delta_{\text{Ped}}} & \Phi_{\delta_{\text{Lat}}} \\ \frac{1}{\tau_e}B_{\delta_{\text{Ped}}} & \frac{1}{\tau_e}B_{\delta_{\text{Lat}}} \end{bmatrix} \quad (59)$$

$$\bar{x}_{\text{long-ver}} = \begin{bmatrix} u & w & q & \theta & a_{1s} \end{bmatrix}^T \quad (60)$$
$$\bar{x}_{\text{lat-dir}} = \begin{bmatrix} v & p & r & \phi & b_{1s} \end{bmatrix}^T$$

$$\bar{u}_{\text{long-ver}} = \begin{bmatrix} \delta_{\text{Coll}} & \delta_{\text{Long}} \end{bmatrix}^T \quad (61)$$
$$\bar{u}_{\text{lat-dir}} = \begin{bmatrix} \delta_{\text{Ped}} & \delta_{\text{Lat}} \end{bmatrix}^T$$

### 3.1 Hover dynamics

The hover regime is characterized by weak coupling between longitudinal and lateral directional mode at least for the purpose of instruction. The longitudinal hover dynamics of the X-cell helicopter is represented by the following system and input matrices.

$$\mathbf{A}_{\text{long-ver}} = \begin{bmatrix} -0.0352 & 0 & 0.9953 & -9.8066 & -9.9532 \\ 0 & -0.096 & 0 & 0.0161 & 0.0161 \\ 0.0693 & 0 & -21.5235 & 0 & 102.4125 \\ 0 & 0 & 0.997 & 0 & 0 \\ 0.0032 & 0 & -1 & 0 & -10 \end{bmatrix} \quad (62)$$

$$\mathbf{B}_{\text{long-ver}} = \begin{bmatrix} -0.2063 & -41.8033 \\ 124.4615 & 0 \\ 1.1691 & 903.9850 \\ 0 & 0 \\ 0 & 42 \end{bmatrix}$$

The lateral directional hover dynamics of the X-cell helicopter is represented by the following system and input matrices

$$\mathbf{A}_{\text{lat-dir}} = \begin{bmatrix} -0.0698 & -0.9953 & 0 & 9.7771 & -9.9529 \\ 0.1212 & -40.6551 & 0 & 0 & 193.4487 \\ 0.0708 & 0 & 0.0001 & 0 & 0 \\ 0 & 1 & -0.0016 & 0 & 0 \\ 0.0032 & -1 & 0 & 0 & -10 \end{bmatrix} \quad (63)$$

$$\mathbf{B}_{\text{lat-dir}} = \begin{bmatrix} -4.7246 & 41.8026 \\ -17.2186 & 1707.5153 \\ 125.9109 & 0 \\ 0 & 0 \\ 0 & 42 \end{bmatrix}$$

The corresponding eigen values along with their frequency and damping for the system matrix are

| Eigen value | Damping ratio | Frequency (rad/s) |
|---|---|---|
| $-9.6 \times 10^{-2}$ | 1 | $9.6 \times 10^{-2}$ |
| $-1.55 \times 10^{-2} \pm 1.77 \times 10^{-1}i$ | $8.74 \times 10^{-2}$ | $1.77 \times 10^{-1}$ |
| $-15.8 \pm 8.32i$ | $8.84 \times 10^{-1}$ | 17.8 |

**Table 10: Hover, Longitudinal**

| Eigen value | Damping ratio | Frequency (rad/s) |
|---|---|---|
| $4.54 \times 10^{-2}$ | -1 | $4.54 \times 10^{-2}$ |
| $1.08 \times 10^{-1}$ | -1 | $1.08 \times 10^{-1}$ |
| $-2.27 \times 10^{-1}$ | 1 | $2.27 \times 10^{-1}$ |
| $-18.9$ | 1 | 18.9 |
| $-31.8$ | 1 | 31.8 |

**Table 11: Hover, Lateral-Directional**

There are two eigen values with positive real parts located close to the origin. They show that the helicopter during hover is marginally unstable

To analyze the dominant contribution within each mode, we can look at the plot of the eigen vectors. First, we need to normalize each state to properly determine its relative dominance as follows.

$$\hat{u} = \frac{u}{\Omega R}, \hat{w} = \frac{w}{\Omega R}, \hat{q} = \frac{q}{\Omega} \quad (64)$$

$$\hat{v} = \frac{v}{\Omega R}, \hat{p} = \frac{p}{R}, \hat{r} = \frac{r}{\Omega} \quad (65)$$

The results of the comparison are tabulated as the following:

|  | Short period | Phugoid | Heaving |
|---|---|---|---|
| $\dot{u}$ | $-4.62 \times 10^{-4} + 2.5 \times 10^{-5}i$ | $7.73 \times 10^{-3}$ | $2.3 \times 10^{-8}$ |
| $\dot{w}$ | $1.4 \times 10^{-7} - 3.6 \times 10^{-7}i$ | $-1.094 \times 10^{-5} - 3.8 \times 10^{-6}i$ | $-7.73 \times 10^{-3}$ |
| $\dot{q}$ | $5.94 \times 10^{-3}$ | $1.93 \times 10^{-5} - 1.00 \times 10^{-7}i$ | $-1.4 \times 10^{-11}$ |
| $\dot{\theta}$ | $-4.9 \times 10^{-2} - 2.6 \times 10^{-2}i$ | $-1.68 \times 10^{-3} - 1.8 \times 10^{-2}i$ | $2.4 \times 10^{-8}$ |
| $\dot{a}_{1s}$ | $5.58 \times 10^{-2} + 8.06 \times 10^{-2}i$ | $-1.3 \times 10^{-7} + 1.8 \times 10^{-6}i$ | $1.2 \times 10^{-9}$ |

**Table 12**

|  | 1st | 2nd | 3rd | 4th | 5th |
|---|---|---|---|---|---|
| $\dot{v}$ | $4.28 \times 10^{-4}$ | $1.062 \times 10^{-3}$ | $7.37 \times 10^{-3}$ | $-6.44 \times 10^{-3}$ | $4.26 \times 10^{-3}$ |
| $\dot{p}$ | $5.97 \times 10^{-3}$ | $5.89 \times 10^{-3}$ | $1.76 \times 10^{-5}$ | $-1.5 \times 10^{-5}$ | $1.0 \times 10^{-5}$ |
| $\dot{r}$ | $-7.4 \times 10^{-7}$ | $3.1 \times 10^{-6}$ | $-1.781 \times 10^{-3}$ | $-3.31 \times 10^{-3}$ | $5.0 \times 10^{-2}$ |
| $\dot{\phi}$ | $-3.14 \times 10^{-2}$ | $5.2 \times 10^{-2}$ | $-1.51 \times 10^{-2}$ | $-1.533 \times 10^{-2}$ | $6.76 \times 10^{-3}$ |
| $\dot{b}_{1s}$ | $4.58 \times 10^{-2}$ | $1.105 \times 10^{-1}$ | $1.45 \times 10^{-5}$ | $-1.4 \times 10^{-5}$ | $9.0 \times 10^{-6}$ |

**Table 13**

The plots of the eigen vectors are presented in the following figures. **Figure 12:** shows that the short period mode is dominated by the pitch angle response $\theta$ and the longitudinal flapping $a_{1s}$. **Figure 13:** shows that the heaving mode is dominated by the vertical velocity $w$. **Figure 14:** indicates that the phugoid mode is composed of forward velocity $u$ and the pitch angle $\theta$.

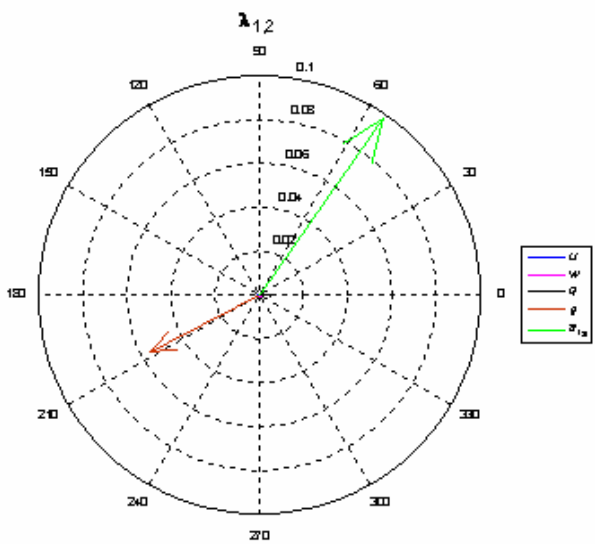

**Figure 12:** Short period mode eigen vectors

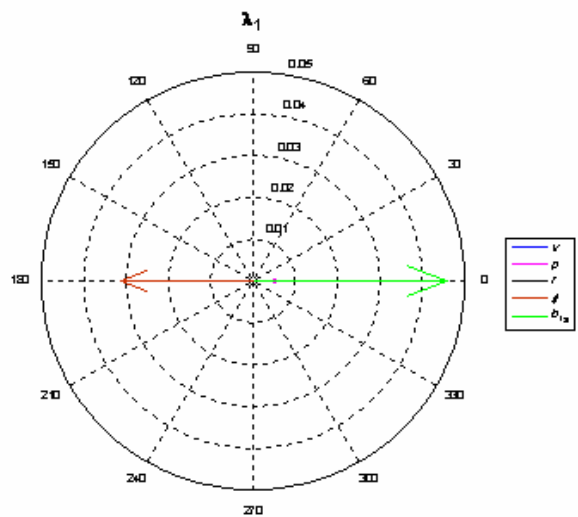

**Figure 15:** 1st mode eigen vectors for hover

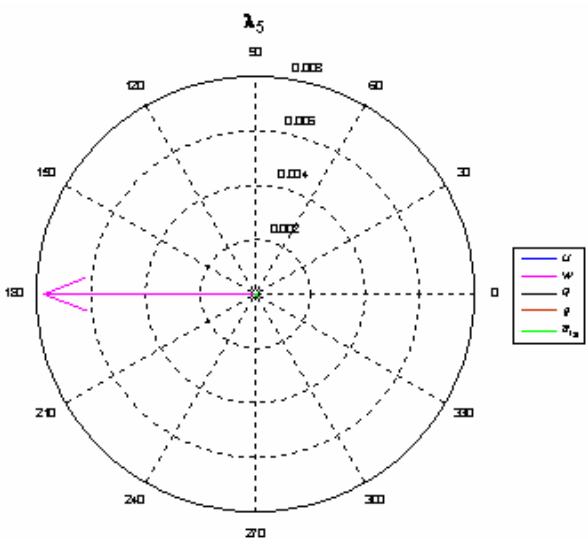

**Figure 13:** Heaving mode eigen vector

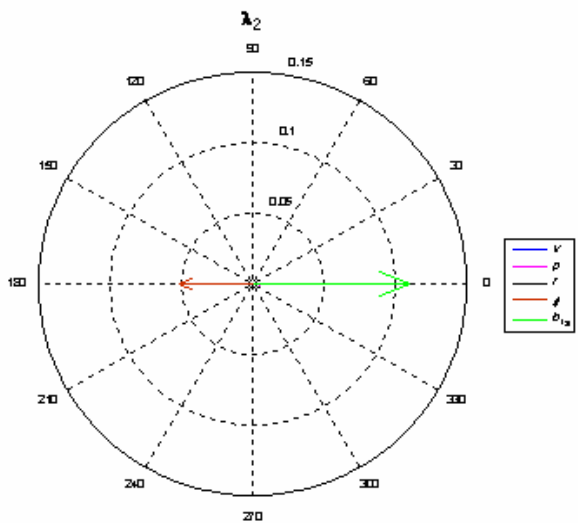

**Figure 16:** 2nd mode eigen vectors for hover

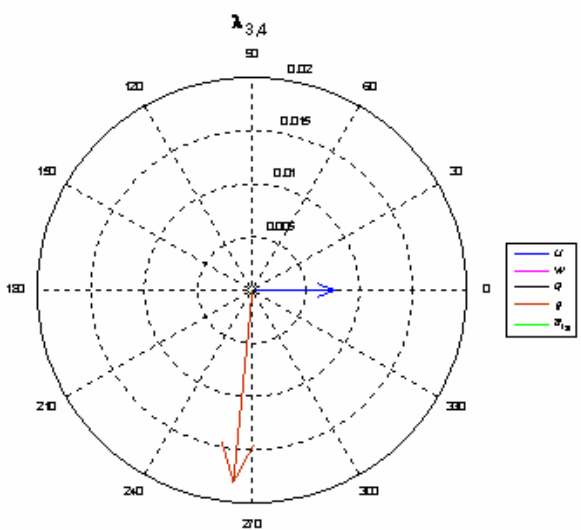

**Figure 14:** Phugoid mode eigen vector

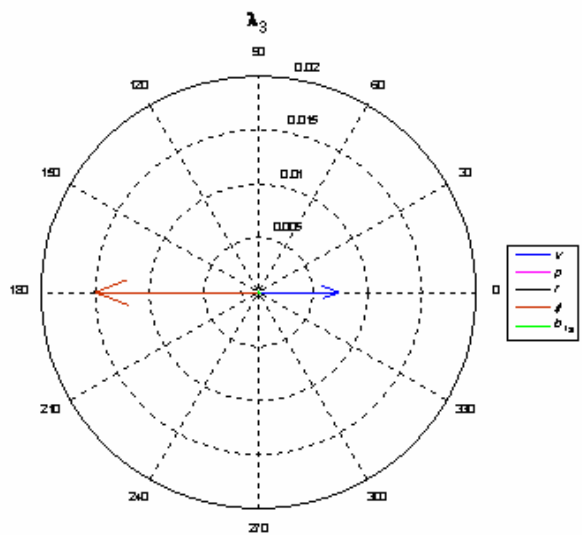

**Figure 17:** 3rd mode eigen vectors for hover

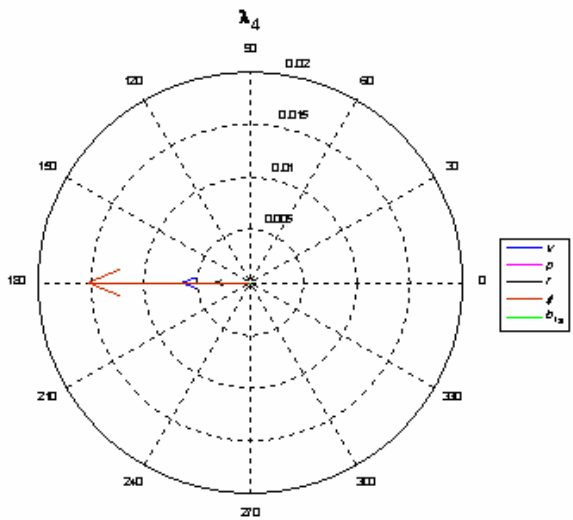

**Figure 18:** 4$^{th}$ mode eigen vectors for hover

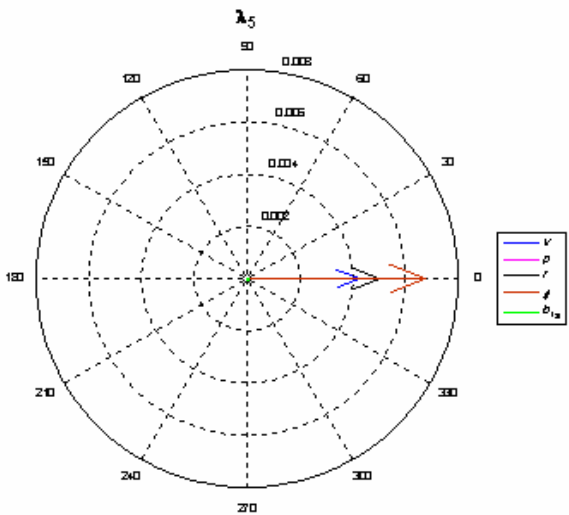

**Figure 19:** 5$^{th}$ mode eigen vectors for hover

### 3.2 Forward flight dynamics

The longitudinal forward flight dynamics of the X-cell helicopter is represented by the following system and input matrices

$$\mathbf{A}_{long\text{-}ver} = \begin{bmatrix} -0.2339 & -0.0072 & 0.7432 & -9.6060 \\ -0.1422 & -1.8927 & 16.5294 & 1.9675 \\ 0.3534 & -0.2983 & -22.0095 & 0 \\ 0 & 0 & 0.9969 & 0 \end{bmatrix}$$

$$\mathbf{B}_{long\text{-}ver} = \begin{bmatrix} -1.6635 & -42.0612 \\ -149.4974 & 0 \\ 37.9729 & 905.2517 \\ 0 & 0 \\ 0 & 42 \end{bmatrix} \quad (66)$$

The lateral-directional forward flight dynamics of the X-cell helicopter is represented by the following state and input matrices.

$$\mathbf{A}_{lat\text{-}dir} = \begin{bmatrix} -0.3468 & -0.7485 & -16.5191 & 9.576 \\ -0.2366 & -40.7435 & 0.1335 & 0 \\ 2.5045 & 0.1406 & -0.976 & 0 \\ 0 & 1 & -0.2048 & 0 \end{bmatrix}$$

$$\mathbf{B}_{lat\text{-}dir} = \begin{bmatrix} -4.8 & 42.1 \\ -17.4 & 1709.9 \\ 127.4 & 0 \\ 0 & 0 \end{bmatrix} \quad (67)$$

In the above expressions, the flapping dynamics is not included. The corresponding eigen values along with their frequency and damping for the system matrix are

| Eigen value | Damping ratio | Frequency (rad/s) |
|---|---|---|
| $-1.16\times10^{-1} \pm 3.70\times10^{-1}i$ | $3.00\times10^{-1}$ | $3.88\times10^{-1}$ |
| -2.12 | 1 | 2.12 |
| -21.8 | 1 | 21.8 |

**Table 14: Forward Flight, Longitudinal**

In general the eigen value shows that the system is stable. The first pair of eigen values represent a lightly damped, long period oscillation which is comparable to phugoid mode in fixed-wing aircraft. The third and fourth eigen value represent a damped pitching mode.

| Eigen value | Damping ratio | Frequency (rad/s) |
|---|---|---|
| $-1.17\times10^{-1}$ | 1 | $1.17\times10^{-1}$ |
| $-6\times10^{-1} \pm 6.42i$ | $9.33\times10^{-2}$ | 6.45 |
| -40.7 | 1 | 40.7 |

**Table 15: Forward Flight, Lateral-Directional**

The lateral-directional eigen value shows that the system is stable. The pair of complex-conjugate eigen values represent a combination of roll and yaw motion corresponding to dutch-roll mode in fixed-wing aircraft. The first eigen value is a stable spiral mode while the fourth represents a roll subsidence.

The eigen vectors analysis is similarly conducted as in the hover case as follows.

| | Short period | Phugoid | Heaving |
|---|---|---|---|
| $\dot{u}$ | $-2.31\times10^{-4}+8.8\times10^{-5}i$ | $7.7\times10^{-3}$ | $1.77\times10^{-4}$ |
| $\dot{w}$ | $-3.6\times10^{-3}+3.57\times10^{-3}i$ | $-1.4\times10^{-4}+1.8\times10^{-4}i$ | $7.7\times10^{-3}$ |
| $\dot{q}$ | $4.48\times10^{-3}$ | $5\times10^{-4}+1.5\times10^{-6}i$ | $-4.16\times10^{-5}$ |
| $\dot{\theta}$ | $-2.8\times10^{-2}+2.44\times10^{-2}i$ | $-1.08\times10^{-2}+2.71\times10^{-2}i$ | $3.5\times10^{-3}$ |
| $\dot{a}_{1s}$ | $2.31\times10^{-2}+4.48\times10^{-2}i$ | $-7.8\times10^{-4}+6.5\times10^{-6}i$ | $1.1\times10^{-3}$ |

**Table 16**

The eigen vector plots are given in the following figures. For cruise condition, the heaving mode consist of combination of $w$ and $\theta$. The short period mode is dominated by $\theta$ and the longitudinal flapping as in the case of hover.

## 4 Conclusions

The linear model developed in the previous chapter is obtained by analytically deriving the stability and control derivatives of the small scale helicopter. The stability analysis was conducted in **3** to analyze the key dynamic characteristics of the vehicle without which it will be impossible to produce a model that is accurate but adequately simple to be practical for control design and analysis. Once the model is developed, a validation can be made by comparing the response of the linear model to that of the nonlinear model for the same input performed in the vicinity of the trimmed condition.

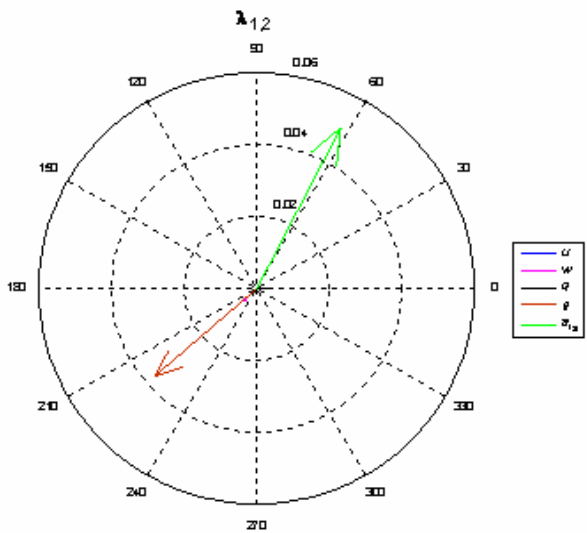

**Figure 20:** Short period mode eigen vectors for cruise

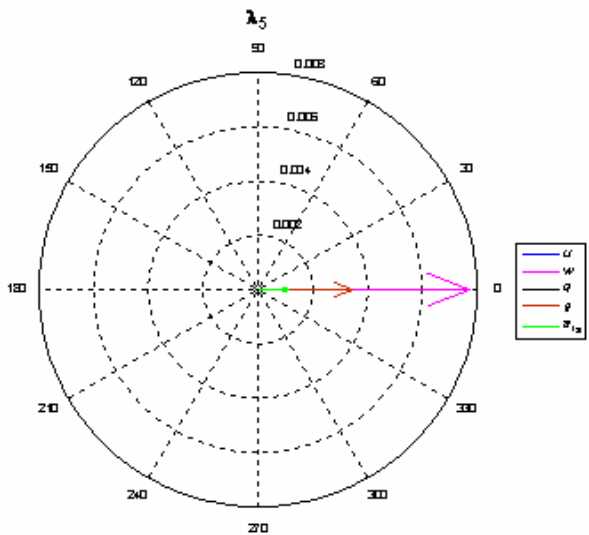

**Figure 21:** Heaving mode eigen vectors

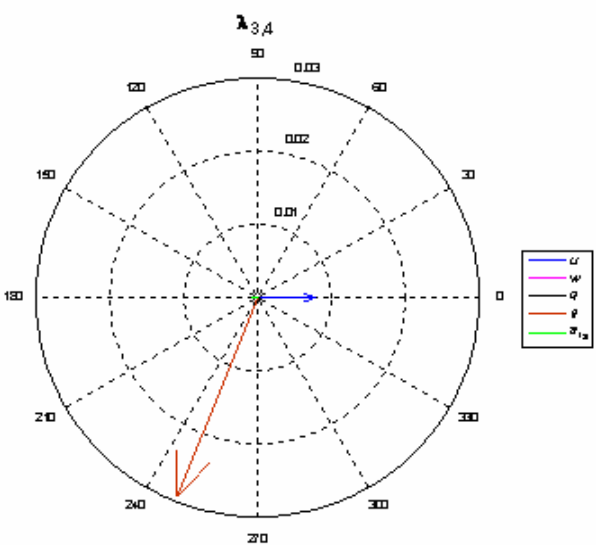

**Figure 22:** Phugoid mode eigen vectors for cruise